\documentclass[twocolumn,compsoc,journal]{IEEEtran}
\usepackage[T1]{fontenc}
\usepackage[latin9]{inputenc}
\usepackage{float}
\usepackage{amsmath}
\usepackage{amssymb}
\usepackage{graphicx}
\usepackage[unicode=true,
 bookmarks=true,bookmarksnumbered=true,bookmarksopen=true,bookmarksopenlevel=1,
 breaklinks=false,pdfborder={0 0 0},pdfborderstyle={},backref=false,colorlinks=false]
 {hyperref}
\hypersetup{pdftitle={Your Title},
 pdfauthor={Your Name},
 pdfpagelayout=OneColumn, pdfnewwindow=true, pdfstartview=XYZ, plainpages=false}

\makeatletter
\usepackage[caption=false,font=normalsize,labelfont=sf,textfont=sf]{subfig}

\makeatother

\begin{document}

\title{Training of Deep Neural Networks based on Distance Measures using
RMSProp}

\author{Thomas~Kurbiel~and~Shahrzad~Khaleghian\IEEEcompsocitemizethanks{\IEEEcompsocthanksitem Thomas~Kurbiel is with the Qiagen GmbH, Hilden,
Germany\protect \\
e-mail: \href{http://Thomas.Kurbiel@qiagen.com}{Thomas.Kurbiel@qiagen.com}.

\IEEEcompsocthanksitem Shahrzad~Khaleghian is with the Department
of Marketing,\protect \\
Mercator School of Management, University Duisburg-Essen, Germany\protect 

e-mail: \href{http://Shahrzad.Kurbiel@uni-due.de}{Shahrzad.Kurbiel@uni-due.de}.

}}

\IEEEtitleabstractindextext{
\begin{abstract}
The vanishing gradient problem was a major obstacle for the success
of deep learning. In recent years it was gradually alleviated through
multiple different techniques. However the problem was not really
overcome in a fundamental way, since it is inherent to neural networks
with activation functions based on dot products. In a series of papers,
we are going to analyze alternative neural network structures which
are not based on dot products. In this first paper, we revisit neural
networks built up of layers based on distance measures and Gaussian
activation functions. These kinds of networks were only sparsely used
in the past since they are hard to train when using plain stochastic
gradient descent methods. We show that by using Root Mean Square Propagation
(RMSProp) it is possible to efficiently learn multi-layer neural networks.
Furthermore we show that when appropriately initialized these kinds
of neural networks suffer much less from the vanishing and exploding
gradient problem than traditional neural networks even for deep networks.
\end{abstract}


}
\maketitle

\section{Introduction}

Most types of neural networks nowadays are trained using stochastic
gradient descent (SGD). However gradient-based approaches (batched
and stochastic) suffer from several drawbacks. The most critical drawback
is the vanishing gradient problem, which makes it hard to learn the
parameters of the \textquotedbl{}front\textquotedbl{} layers in an
$n$-layer network. This problem becomes worse as the number $n$
of layers in the architecture increases. This is especially critical
in recurrent neural networks \cite{Hochreiter1997}, which are trained
by unfolding them into very deep feedforward networks. For each time
step of an input sequence processed by the network a new layer is
created. The second drawback of gradient-based approaches is the potential
to get stuck at local minima and saddle points.

To overcome the vanishing gradient problem, several methods were proposed
in the past \cite{Glorot2010}\cite{Hahnloser2000}\cite{Hinton2006}\cite{Hochreiter1997}.
Early methods consisted of the pre-training of the weights using unsupervised
learning techniques in order to establish a better initial configuration,
followed by a supervised fine-tuning through backpropagation. The
unsupervised pre-training was used to learn generally useful feature
detectors. At first, Hinton et. al. used stacked Restricted Boltzmann
Machines (RBM) to do a greedy layer-wise unsupervised training of
deep neural networks \cite{Hinton2006}.

The Xavier Initialization \cite{Glorot2010} by Glorot et. al. (after
Glorot\textquoteright s first name) was a further step to overcome
the vanishing gradient problem. In Xavier initialization the initial
weights of the neural networks are sampled from a Gaussian distribution
where the variance is a function of the number of neurons in a given
layer. At about the same time Rectified Linear Units (ReLU) where
introduced, a non-linearity that is scale-invariant around 0 and does
not saturate at large input values \cite{Hahnloser2000}. Both Xavier
Initialization and ReLUs alleviated the problems the sigmoid activation
functions had with vanishing gradient.

Another method particularly used to cope with the vanishing gradient
problem for recurrent neural networks is the Long Short-Term Memory
(LSTM) network \cite{Hochreiter1997}. By introducing memory cells
the gradient can be propagated back to a much earlier time without
vanishing.

One of the newest and most effective ways to resolve the vanishing
gradient problem is with residual neural networks (ResNets) \cite{2015arXiv151203385H}.
ResNets yield lower training error (and test error) than their shallower
counterparts simply by reintroducing outputs from shallower layers
in the network to compensate for the vanishing data.

The main reason behind vanishing gradient problem is the fundamental
difference between the forward and the backward pass in neural networks
with activation functions based on dot products \cite{Bishop2005}.
While the forward pass is highly non-linear in nature, the backward
pass is completely linear due to the fixed values of the activations
obtained in the forward pass. During the backpropagation the network
behaves as a linear system and suffers from the problem of linear
systems, which is the tendency to either explode or die when iterated. 

In this paper, we address the problem of vanishing gradient by exploring
the properties of neural networks based on distance measures which
are an alternative to neural networks based on dot products \cite{Debes2005}.
In neural networks based on distance measures the weight vectors of
neurons represent landmarks in the input space. The activation of
a neuron is computed from the weighted distance of the the input vector
from the corresponding landmark. Determining the distance is possible
by applying different measures. In this work, we apply the most commonly
used Gaussian kernel \cite{Debes2005}. The parameters of the network
to be learnt consist of both the landmarks and the weights of the
distance measure.

\pagebreak

\section{Basic Principle}

\begin{figure}[tbh]
\includegraphics[bb=0bp 10bp 891bp 242bp,scale=0.28]{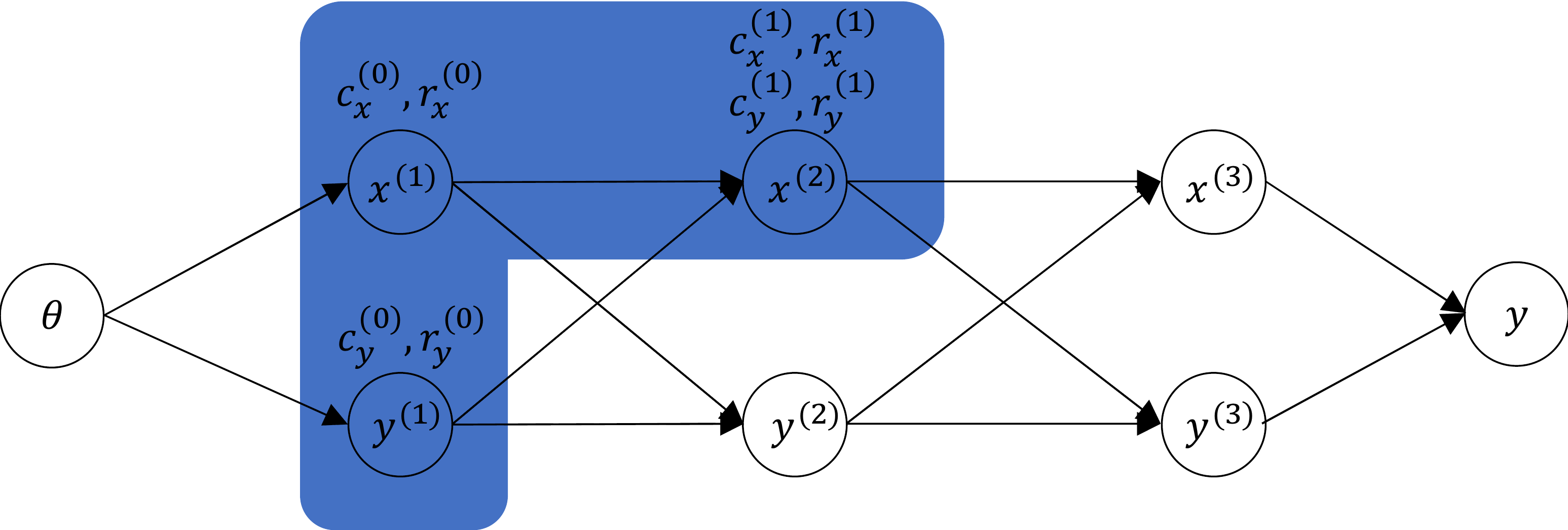}

\caption{Network of Gaussian layers interconnected in a feed-forward way.\label{fig:Network-of-Gaussian}}
\end{figure}

In this section, we show how a network of Gaussian layers interconnected
in a feed-forward way can be used to approximate arbitrary bounded
continuous functions. For demonstration purposes, we use a simple
1D example, see Fig.\ \ref{fig:Network-of-Gaussian}.

\begin{figure}[tbh]
\begin{centering}
\includegraphics[bb=0bp 60bp 806bp 458bp,scale=0.25]{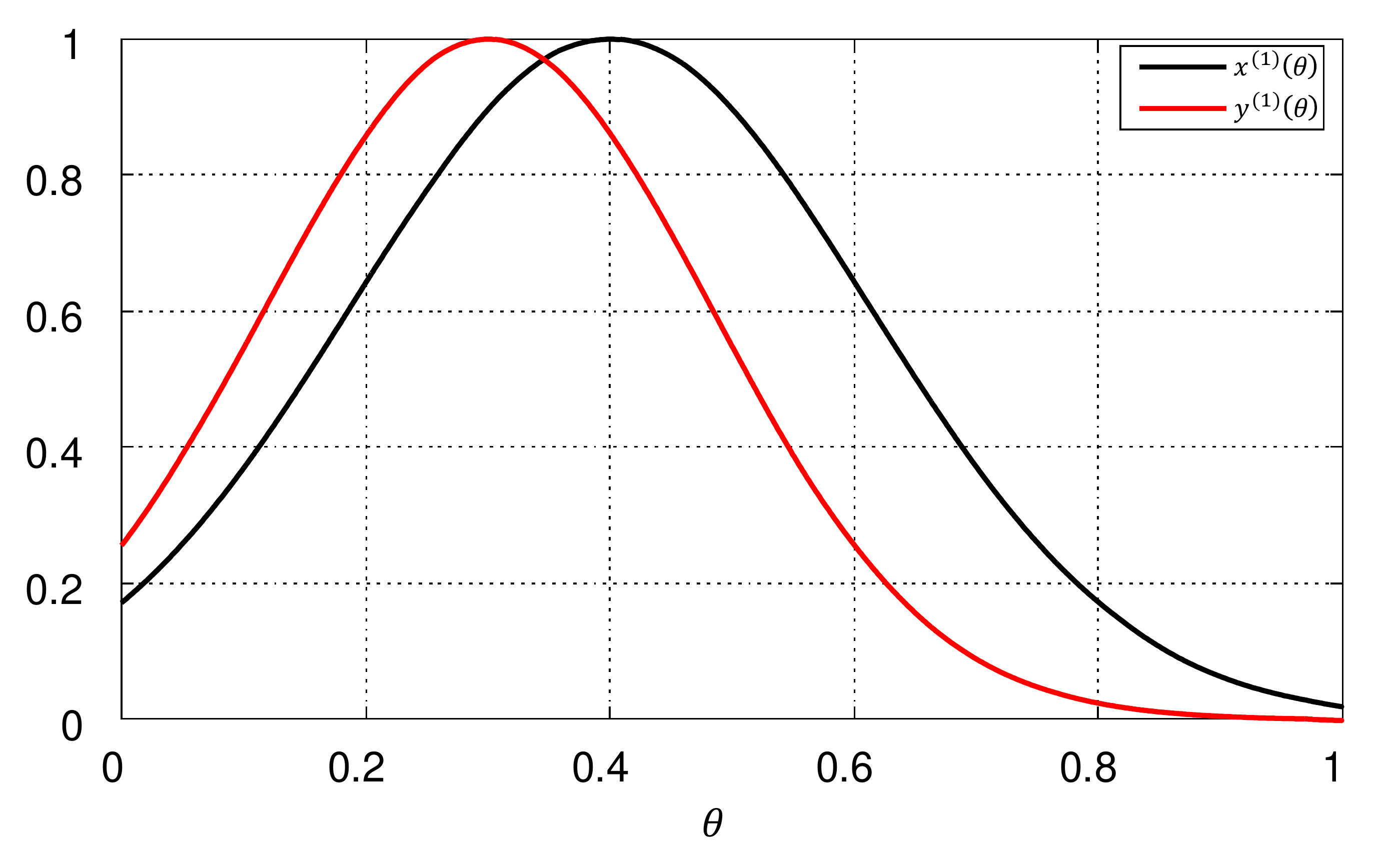}
\par\end{centering}
\centering{}\caption{Activations $x^{\left(1\right)}\left(\theta\right)$ and $y^{\left(1\right)}\left(\theta\right)$
of the first hidden layer. \newline$\;$Used parameters: $r_{\mathrm{x}}^{\left(0\right)}=4.7,\ c_{\mathrm{x}}^{\left(1\right)}=0.4,\ r_{\mathrm{y}}^{\left(1\right)}=5.5,\ c_{\mathrm{y}}^{\left(1\right)}=0.3$\label{fig:First-layer-Used}}
\end{figure}

The input of the network is the variable $\theta\in\mathbb{R}$ restricted
to the interval $\left[0,1\right]$. The first layer of the network
consists of two simple 1D-Gaussian functions:
\begin{equation}
x^{\left(1\right)}\left(\theta\right)=\exp\left(-\frac{1}{2}\cdot\left[r_{\mathrm{x}}^{\left(0\right)}\right]^{2}\cdot\left(\theta-c_{\mathrm{x}}^{\left(0\right)}\right)^{2}\right),\label{eq:x_1}
\end{equation}
and
\begin{equation}
y^{\left(1\right)}\left(\theta\right)=\exp\left(-\frac{1}{2}\cdot\left[r_{\mathrm{y}}^{\left(0\right)}\right]^{2}\cdot\left(\theta-c_{\mathrm{y}}^{\left(0\right)}\right)^{2}\right),\label{eq:y_1}
\end{equation}
with $r_{\mathrm{x}}^{\left(0\right)},c_{\mathrm{x}}^{\left(0\right)},r_{\mathrm{y}}^{\left(1\right)},r_{\mathrm{y}}^{\left(0\right)}\in\mathbb{R}$.
The shapes of (\ref{eq:x_1}) and (\ref{eq:y_1}) as functions of
$\theta$ are depicted in Fig.\ \ref{fig:First-layer-Used}.

\begin{figure}[tbh]
\begin{centering}
\includegraphics[bb=0bp 40bp 599bp 458bp,scale=0.29]{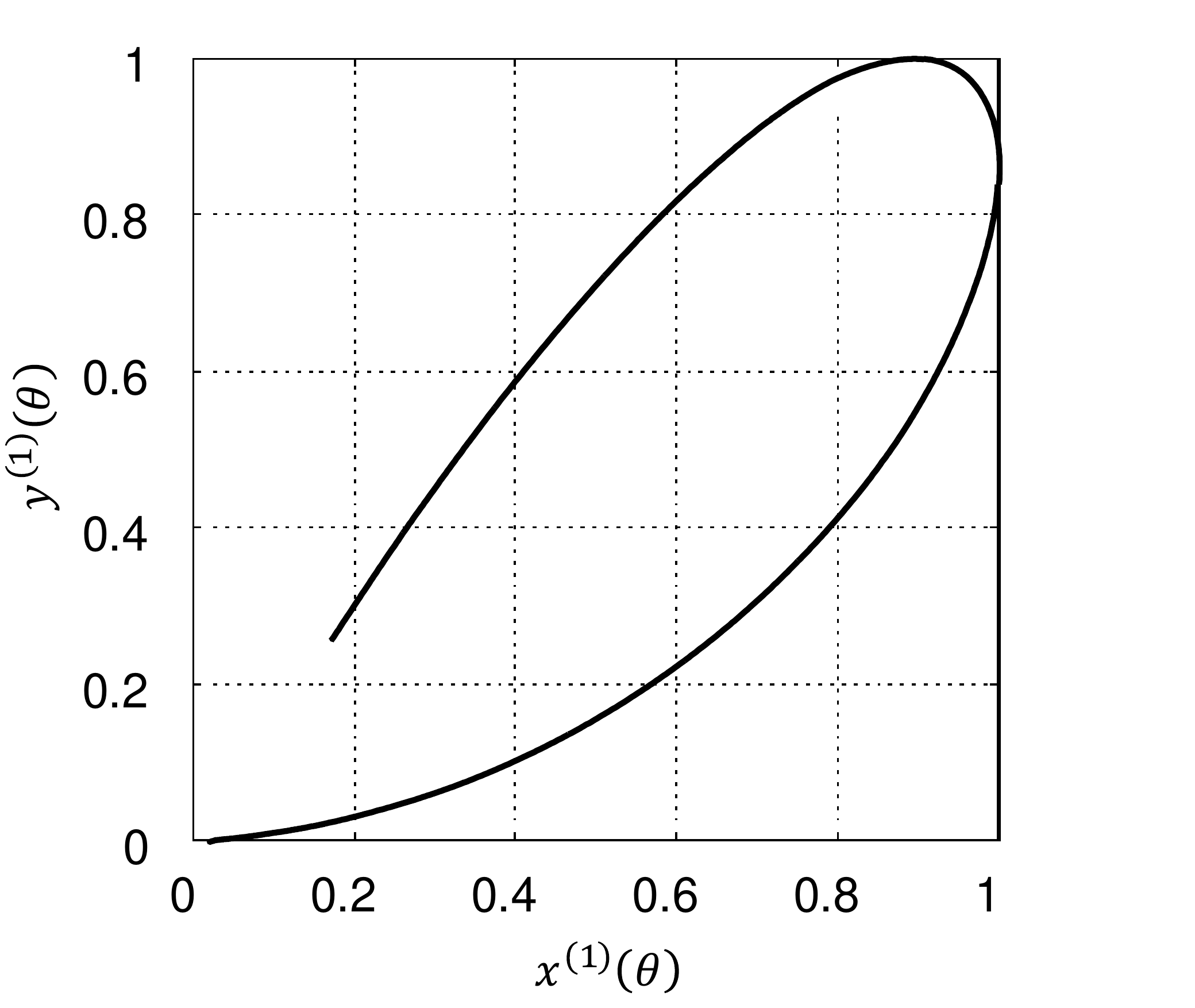}
\par\end{centering}
\centering{}\caption{2D trajectory composed of two 1D-Gaussians.\label{fig:2D-trajectory-composed}}
\end{figure}

The two outputs of the first layer can be interpreted as the x-coordinate
and the y-coordinate of a 2D trajectory. The resulting trajectory
is depicted in Fig.\ \ref{fig:2D-trajectory-composed}.

\begin{figure}[htbp]
\begin{centering}
\includegraphics[bb=30bp 80bp 797bp 417bp,scale=0.35]{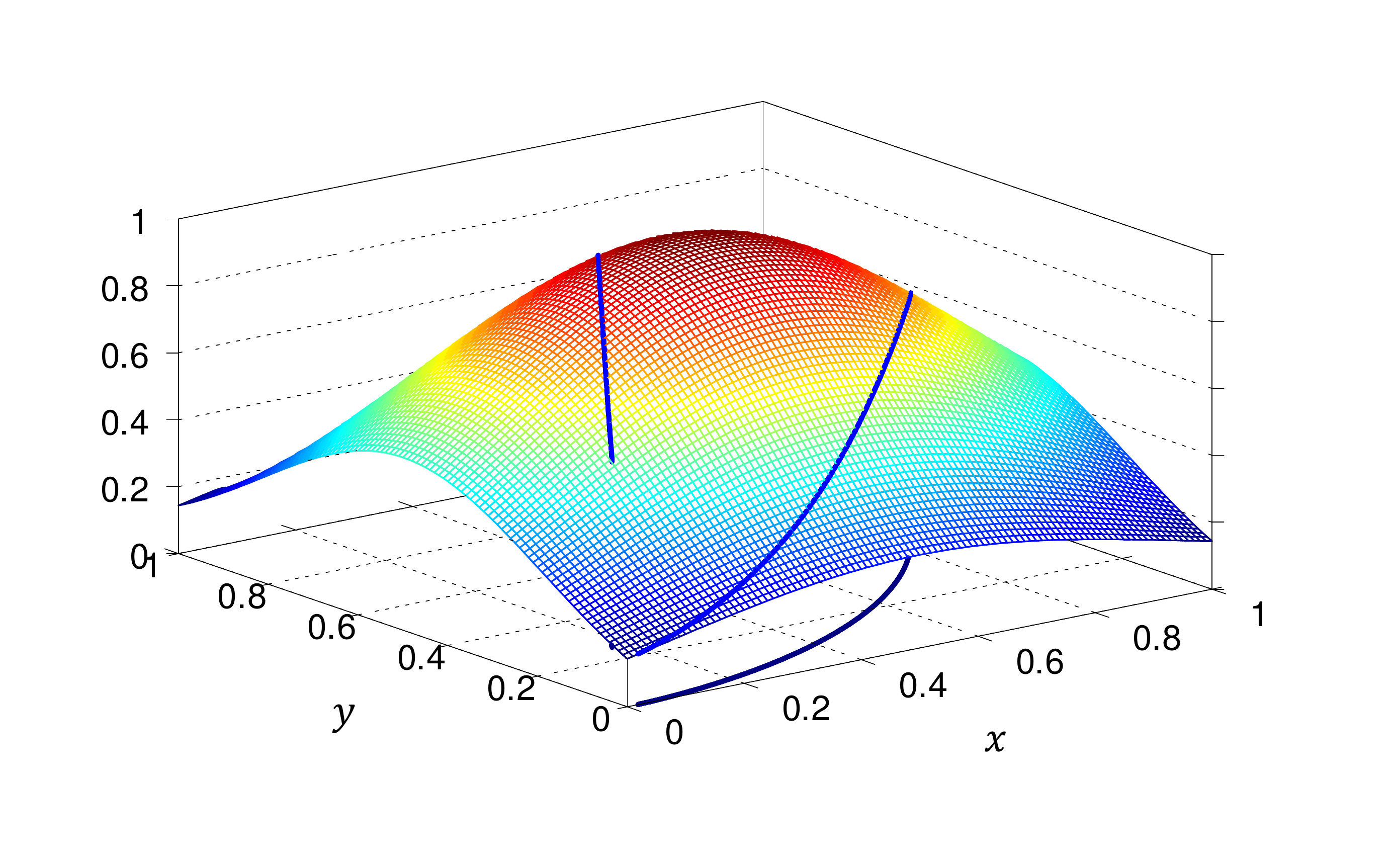}
\par\end{centering}
\centering{}\caption{Trajectory generated by traverse of a 2D-Gaussian along a path. \newline
Used parameters: $r_{\mathrm{x}}^{\left(1\right)}=2.3,\ c_{\mathrm{x}}^{\left(1\right)}=0.5,\ r_{\mathrm{y}}^{\left(1\right)}=3.2,\ c_{\mathrm{y}}^{\left(1\right)}=0.5$\label{fig:Trajectory-generated-by}}
\end{figure}

The resulting trajectory is used as input to the second layer consisting
of two 2D-Gaussian functions, see Fig.\ \ref{fig:Network-of-Gaussian}.
The upper unit is defined as follows:
\begin{equation}
\begin{array}{ccc}
x^{\left(2\right)}\left(\theta\right) & = & \exp(-\frac{1}{2}\cdot\left[r_{\mathrm{x}}^{\left(1\right)}\right]^{2}\cdot\left(x^{\left(1\right)}\left(\theta\right)-c_{\mathrm{x}}^{\left(1\right)}\right)^{2}\\
 &  & \phantom{\exp(}-\frac{1}{2}\cdot\left[r_{\mathrm{y}}^{\left(1\right)}\right]^{2}\cdot\left(y^{\left(1\right)}\left(\theta\right)-c_{\mathrm{y}}^{\left(1\right)}\right)^{2},
\end{array}
\end{equation}
please note that the used 2D-Gaussian function is aligned with both
axes. In particular, no rotation matrix has to be used. Next we traverse
the 2D-Gaussian function along the trajectory defined by (\ref{eq:x_1})
and (\ref{eq:y_1}). The resulting trajectory in 3D space is depicted
in Fig.\ \ref{fig:Trajectory-generated-by}. 

\begin{figure}[tbh]
\begin{centering}
\includegraphics[bb=0bp 60bp 840bp 458bp,scale=0.25]{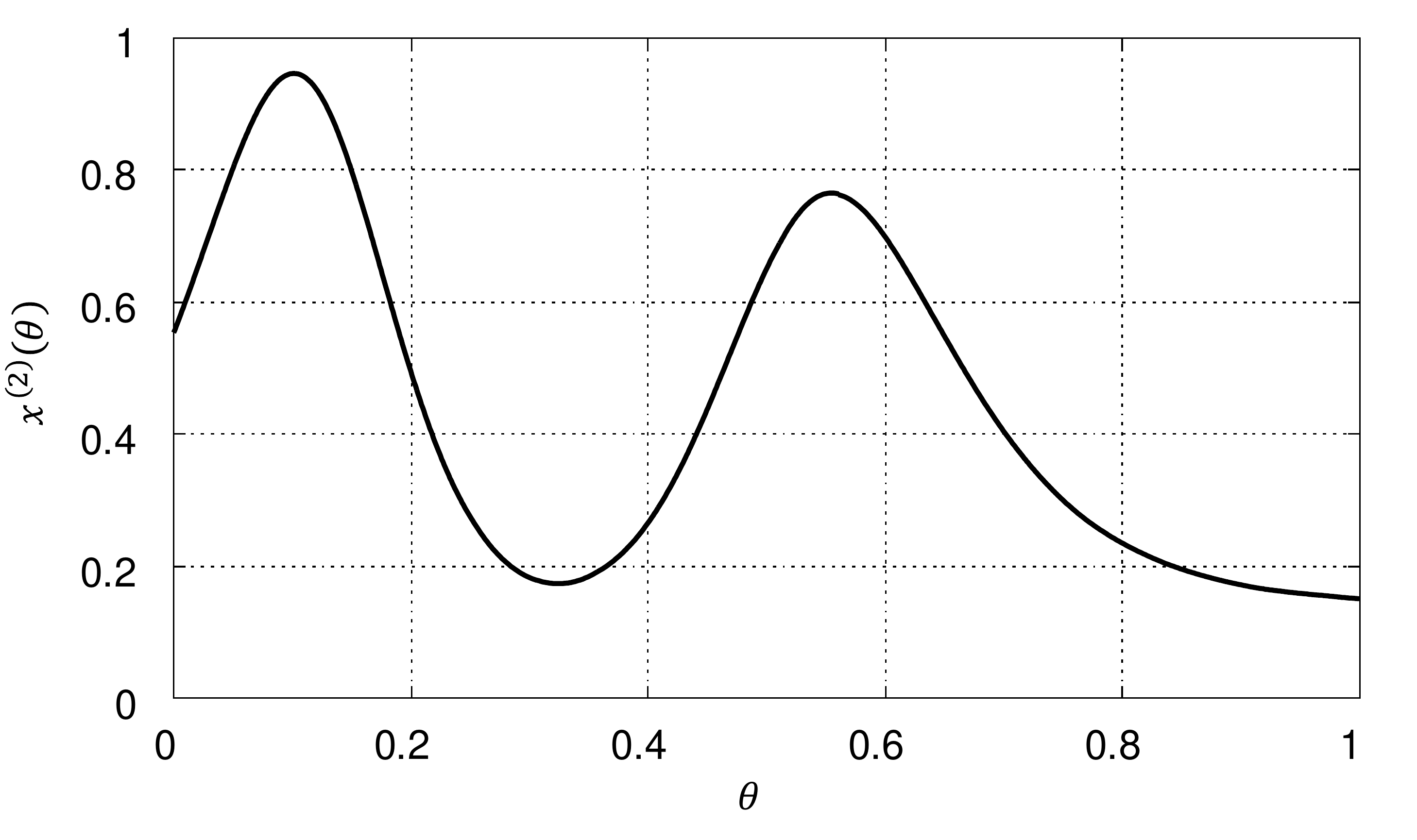}
\par\end{centering}
\caption{Activation $x^{\left(2\right)}\left(\theta\right)$ of the second
hidden layer as a function of $\theta$.\label{fig:Resultung_Trajectory} }
\end{figure}

If we track a point traversing the trajectory, we see that it first
climbs the 2D Gaussian function, then descends and then climbs again.
Hence the resulting curve as a function of $\theta$ exhibits a higher
complexity than each of the constituent parts, see Fig.\ \ref{fig:Resultung_Trajectory}.

A curve generated in the described manner can now be used to define
e.g. the x-coordinate of a much more complex path than in Fig.\ \ref{fig:2D-trajectory-composed}.
Repeating this approach in a recursive manner allows to approximate
continuous functions as demonstrated below.

\begin{figure}[tbh]
\begin{centering}
\includegraphics[bb=0bp 60bp 840bp 458bp,scale=0.25]{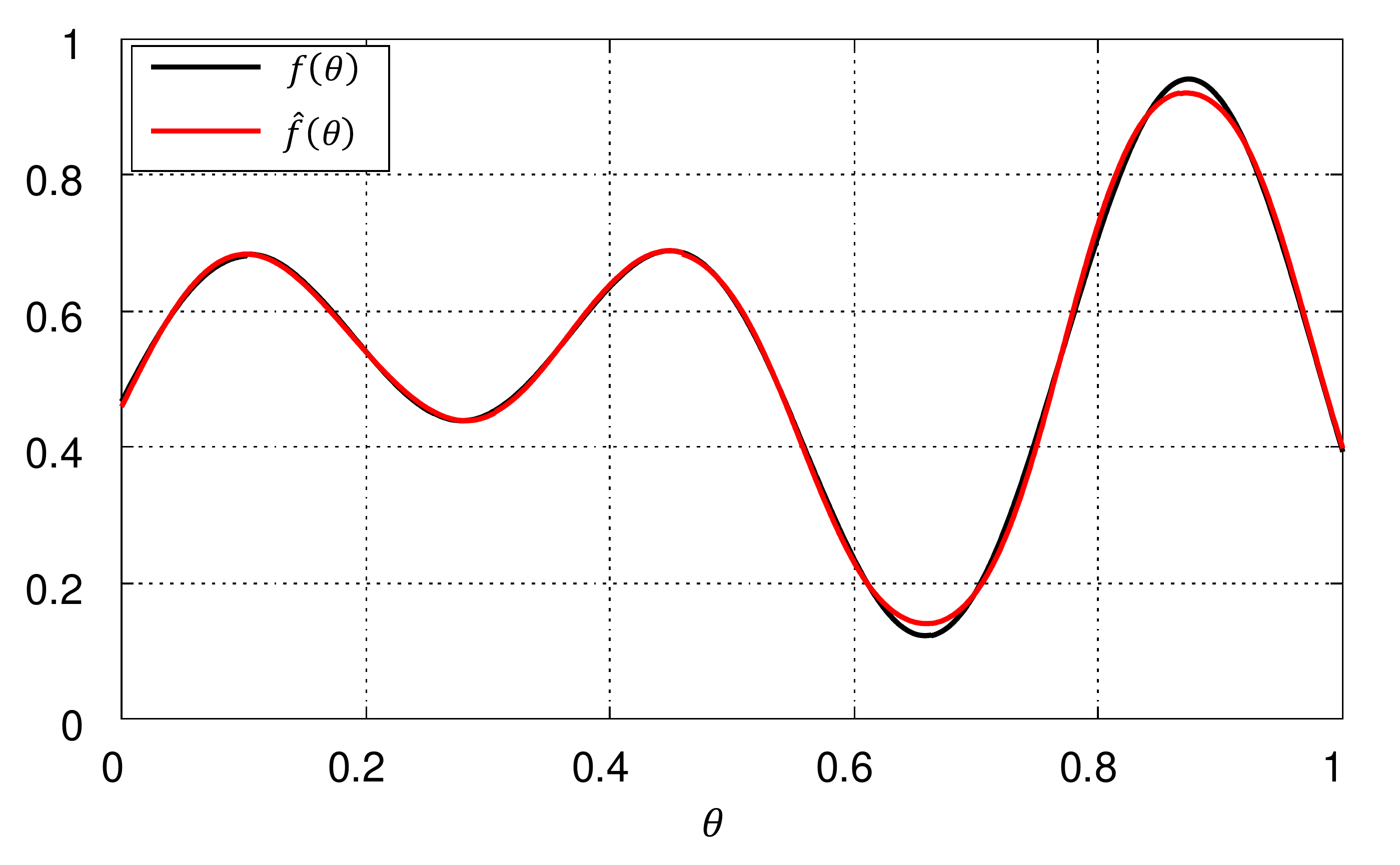}
\par\end{centering}
\caption{Approximation of a 1D-function using a 3-hidden layer network\label{fig:Function_to_be_Approximated} }
\end{figure}

To show the strength of the described method, next we are going to
approximate the following function using the simple three layer network
depicted in Fig.\ \ref{fig:Network-of-Gaussian}:

\begin{equation}
f\left(\theta\right)=\frac{1}{2}+\frac{1}{6}\left(\frac{\sin\left(3\pi\theta-\pi/4\right)}{\exp\left(2/5\cdot\theta\right)}+\frac{\cos\left(5\pi\theta-\pi/3\right)}{\exp\left(-4/5\cdot\theta\right)}\right)\label{eq:Function_to_be_Approximated}
\end{equation}
The shape of (\ref{eq:Function_to_be_Approximated}) together with
its approximation is depicted in Fig.\ \ref{fig:Function_to_be_Approximated}.
The parameters of the network are optimized using the backpropagation
algorithm introduced in the next section. The (Root Mean Square) RMS
of the approximation is 0.008.

The described method can be extended to an arbitrary number of input
and output dimensions.

\section{Forward Pass$\protect\phantom{}$}

\begin{figure}[tbh]
\begin{centering}
\includegraphics[bb=0bp 10bp 455bp 367bp,scale=0.33]{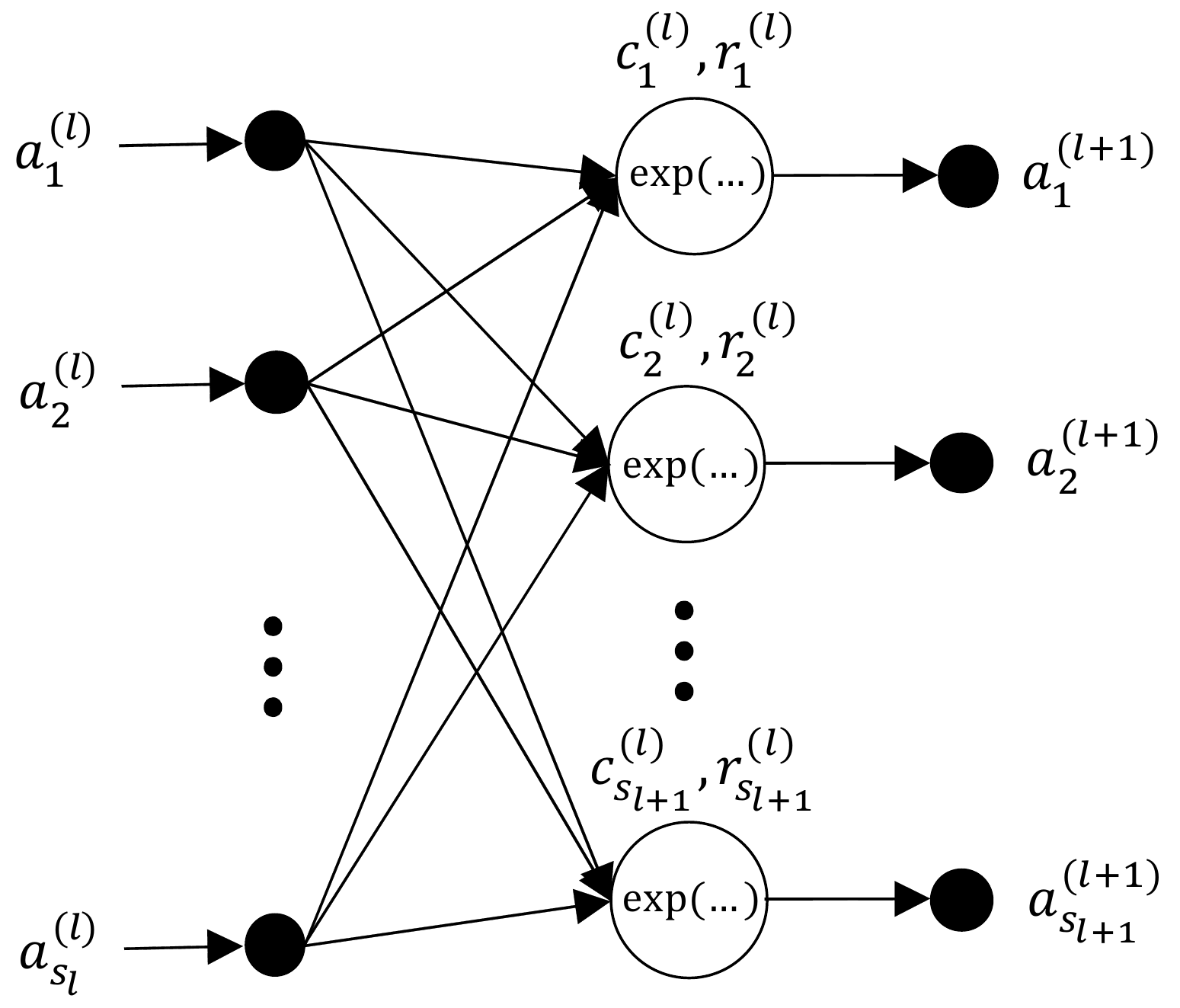}
\par\end{centering}
\caption{Architecture of layer $l$ and corresponding parameters\label{fig:Architecture-of-layer}}
\end{figure}

Layer $l$ has $s_{l}$ inputs denoted by $a_{i}^{\left(l\right)}$
and $s_{l+1}$ outputs denoted by $a_{j}^{\left(l+1\right)}$, as
depicted in Fig.\ \ref{fig:Architecture-of-layer}. The value of
$j$-th output depends on all $s_{l}$ inputs of layer $l$ and is
calculated using the following Gaussian function:
\begin{equation}
a_{j}^{\left(l+1\right)}=\exp\left(-\frac{1}{2}\cdot\sum_{i=1}^{s_{l}}\left[r_{j}^{\left(l\right)}\right]_{i}^{2}\cdot\left(a_{i}^{\left(l\right)}-\left[c_{j}^{\left(l\right)}\right]_{i}\right)^{2}\right),\label{eq:Input_Output_Layer_l}
\end{equation}
where $j\in\left\{ 1,\ldots,s_{l+1}\right\} $. The terms $\left[r_{j}^{\left(l\right)}\right]_{i}$
and $\left[c_{j}^{\left(l\right)}\right]_{i}$ denote the $i$-th
element of vector $r_{j}^{\left(l\right)}\in\mathbb{R}^{s_{l}}$ and
$c_{j}^{\left(l\right)}\in\mathbb{R}^{s_{l}}$ respectively\textit{}\footnote{\textit{By multiplying out the square term in Eq.\ (\ref{eq:Input_Output_Layer_l})
and splitting it up into 3 terms it is possible to organize all parameters
in matrices and use matrix-vector operations. Thus one can take advantage
of fast linear algebra routines to quickly perform calculations in
the network.}}.

\section{Cost function$\protect\phantom{}$}

A cost function measures how well the neural network performs to map
training samples $x^{\left(i\right)}$ to the desired output $y^{\left(i\right)}$.
We denote the cost function with respect to a single training sample
to be:
\begin{equation}
J\left(R,C;\ x^{\left(i\right)},y^{\left(i\right)}\right)\label{eq:individual training examples}
\end{equation}
where $R$ and $C$ represent the parameters of the network based
on distance measures. Given a training set of $m$ samples, the overall
cost function is an average over the cost functions (\ref{eq:individual training examples})
for individual training examples plus a regularization term \cite{Goodfellow-et-al-2016}.
All cost functions typically used in practice with neural networks
with activation functions based on dot products can be applied as
well. 

\begin{figure}[tbh]
\begin{centering}
\includegraphics[scale=0.3]{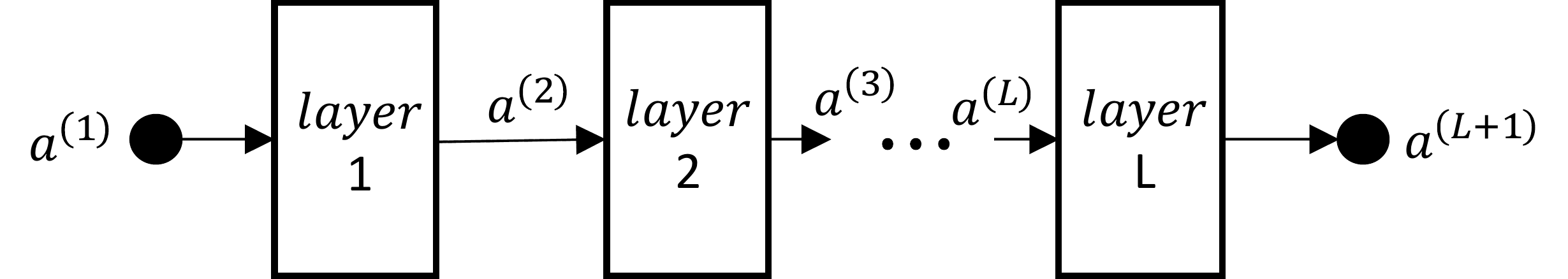}
\par\end{centering}
\caption{Input and output activation functions of a network with $L$ layers\label{fig:Naming-convention}}
\end{figure}

The activations of the output layer are $a_{j}^{\left(L+1\right)},\ j\in\left\{ 1,\ldots,s_{L+1}\right\} $,
where $L$ denotes the number of layers in the network, see Fig.\ \ref{fig:Naming-convention}.
The output activations depends both on the input $x^{\left(i\right)}$
and the parameters $R$ and $C$ of the network: $a_{j}^{\left(L+1\right)}(R,C;\ x^{\left(i\right)})$,
which is implicitly assumed in following formulas. 

In particular for regression tasks we are using the quadratic cost:
\begin{equation}
J\left(R,C;\ x^{\left(i\right)},y^{\left(i\right)}\right)=\sum_{j=1}^{s_{L+1}}\left(a_{j}^{\left(L+1\right)}-y_{j}^{\left(i\right)}\right)^{2},\label{eq:Quadratic_cost}
\end{equation}
where $x^{\left(i\right)}\in\mathbb{R}^{s_{1}}$ is the input vector
and $y^{\left(i\right)}\in\mathbb{R}^{s_{L+1}}$ is the continuous
target value. For classification purposes, we use a Softmax output
layer and the cross-entropy function \cite{Boer2005}:
\begin{equation}
J\left(R,C;\ x^{\left(i\right)},y^{\left(i\right)}\right)=\sum_{j=1}^{s_{L+1}}y_{j}^{\left(i\right)}\log\left(\frac{\exp\ a_{j}^{\left(L+1\right)}}{\sum_{k=1}^{s_{L+1}}\exp\ a_{k}^{\left(L+1\right)}}\right),\label{eq:cross_entropy}
\end{equation}
where $y^{\left(i\right)}$ is a one-hot encoded target output vector.
In case a Softmax output layer is used, the Gaussian activation function
in last layer is omitted:
\begin{equation}
a_{j}^{\left(L+1\right)}=-\frac{1}{2}\cdot\sum_{i=1}^{s_{L}}\left[r_{j}^{\left(L\right)}\right]_{i}^{2}\cdot\left(a_{i}^{\left(L\right)}-\left[c_{j}^{\left(L\right)}\right]_{i}\right)^{2},
\end{equation}

\section{Backward pass$\protect\phantom{}$}

As in conventional neural networks backpropagation is easily adaptable
\cite{Rumelhart1986}\cite{Schmidhuber2015}. Let us denote the partial
derivative of the cost function w.r.t the activation function of layer
$l+1$ as $\delta_{j}^{\left(l+1\right)}=\partial J/\partial a_{j}^{\left(l+1\right)}$.

The upper partial derivative of the predecessing layer is backpropagated
using the following formula:
\begin{equation}
\delta_{i}^{\left(l\right)}=-\sum_{j=1}^{s_{l+1}}\delta_{j}^{\left(l+1\right)}\cdot a_{j}^{\left(l+1\right)}\cdot\left[r_{j}^{\left(l\right)}\right]_{i}^{2}\cdot\left(a_{i}^{\left(l\right)}-\left[c_{j}^{\left(l\right)}\right]_{i}\right),\label{eq:Backpropagation_delta}
\end{equation}
where $i\in\left\{ 1,\ldots,s_{l}\right\} $. The update of the centroid
parameters in layer $l$ is calculated via:
\begin{equation}
\frac{\partial J}{\partial\left[c_{j}^{\left(l\right)}\right]_{i}}=\delta_{j}^{\left(l+1\right)}\cdot a_{j}^{\left(l+1\right)}\cdot\left[r_{j}^{\left(l\right)}\right]_{i}^{2}\cdot\left(a_{i}^{\left(l\right)}-\left[c_{j}^{\left(l\right)}\right]_{i}\right),
\end{equation}
where $j\in\left\{ 1,\ldots,s_{l+1}\right\} $ and $i\in\left\{ 1,\ldots,s_{l}\right\} $.
The update of the radius parameters in layer $l$ is :
\begin{equation}
\frac{\partial J}{\partial\left[r_{j}^{\left(l\right)}\right]_{i}}=-\delta_{j}^{\left(l+1\right)}\cdot a_{j}^{\left(l+1\right)}\cdot\left[r_{j}^{\left(l\right)}\right]_{i}\cdot\left(a_{i}^{\left(l\right)}-\left[c_{j}^{\left(l\right)}\right]_{i}\right)^{2}.
\end{equation}

\section{RMSProp$\protect\phantom{}$}

It is not possible to train the neural networks based on distance
measures and Gaussian activations functions using only plain mini-batch
gradient descent or momentum. No or only extremely slow convergence
can be achieved that way. This is the case even for shallow networks.
The solution is to apply RMSProp (Root Mean Square Propagation), which
renders the training possible in the first place \cite{Tieleman2012}. 

In RMSProp, the learning rate is adapted for each of the parameter
vectors $c_{j}^{\left(l\right)}$ and $r_{j}^{\left(l\right)}$, $\forall j\in\left\{ 1,\ldots,s_{l}\right\} $
and $\forall l\in\left\{ 1,\ldots,L\right\} $. The idea is to keep
a moving average of the squared gradients over adjacent mini-batches
for each weight:
\begin{equation}
\begin{aligned} & \bar{v}\left(c_{j}^{\left(l\right)},\ t\right)=\gamma\bar{v}\left(c_{j}^{\left(l\right)},\ t-1\right)+\left(1-\gamma\right)\left(\partial J/\partial c_{j}^{\left(l\right)}\right)^{2}\\
 & \bar{v}\left(r_{j}^{\left(l\right)},\ t\right)=\gamma\bar{v}\left(r_{j}^{\left(l\right)},\ t-1\right)+\left(1-\gamma\right)\left(\partial J/\partial r_{j}^{\left(l\right)}\right)^{2}
\end{aligned}
\end{equation}
where $\gamma$ is the forgetting factor, typically 0.9. Next the
learning rate for a weight is divided by that moving average:
\begin{equation}
\begin{aligned} & c_{j}^{\left(l\right)}:=c_{j}^{\left(l\right)}-\frac{\eta}{\sqrt{\bar{v}\left(c_{j}^{\left(l\right)},\ t\right)}}\cdot\partial J/\partial c_{j}^{\left(l\right)}\\
 & r_{j}^{\left(l\right)}:=r_{j}^{\left(l\right)}-\frac{\eta}{\sqrt{\bar{v}\left(r_{j}^{\left(l\right)},\ t\right)}}\cdot\partial J/\partial r_{j}^{\left(l\right)}
\end{aligned}
\end{equation}
RMSProp has shown excellent adaptation of learning rate in different
applications. 

\section{Initialization$\protect\phantom{}$}

As in neural networks based on dot products \cite{Glorot2010}, a
sensible initialization of the weights is crucial for convergence,
especially when training deep neural networks. The initialization
of the weights controlling the center of the Gaussian functions is
straightforward:

\begin{equation}
\left[c_{j}^{\left(l\right)}\right]_{i}\in\mathcal{N}\left(\mu_{\mathrm{c}},\ \sigma_{\mathrm{c}}^{2}\right),\label{eq:init_centroid}
\end{equation}
where $\mathcal{N}$ denotes a normal distribution. Experiments show
that values $\mu_{\mathrm{c}}=0.63$ and $\sigma_{\mathrm{c}}=0.1$
work well for most network architectures. The initialization in (\ref{eq:init_centroid})
does not depend neither on the number of inputs nor on the number
of outputs of a layer.

The initialization of the weights controlling the radii of the Gaussian
functions is slightly more complicated:
\begin{equation}
\left[r_{j}^{\left(l\right)}\right]_{i}\in\mathcal{N}\left(\mu_{\mathrm{r}},\ \sigma_{\mathrm{r}}^{2}\right),
\end{equation}
with
\begin{equation}
\mu_{\mathrm{r}}=\sqrt{\frac{2\cdot\ln\left(1/\epsilon\right)}{s_{l}}},\ \sigma_{\mathrm{r}}=f_{\mathrm{r}}\cdot\mu_{\mathrm{r}},\label{eq:init_radius}
\end{equation}
where typically $\epsilon=0.1$ and $f_{\mathrm{r}}=0.4$. For breaking
the symmetry setting $f_{\mathrm{r}}>0$ is important. The detailed
derivation of (\ref{eq:init_radius}) will be presented in a separate
paper. 

\begin{figure}[H]
\begin{centering}
\includegraphics[bb=0bp 50bp 797bp 458bp,scale=0.3]{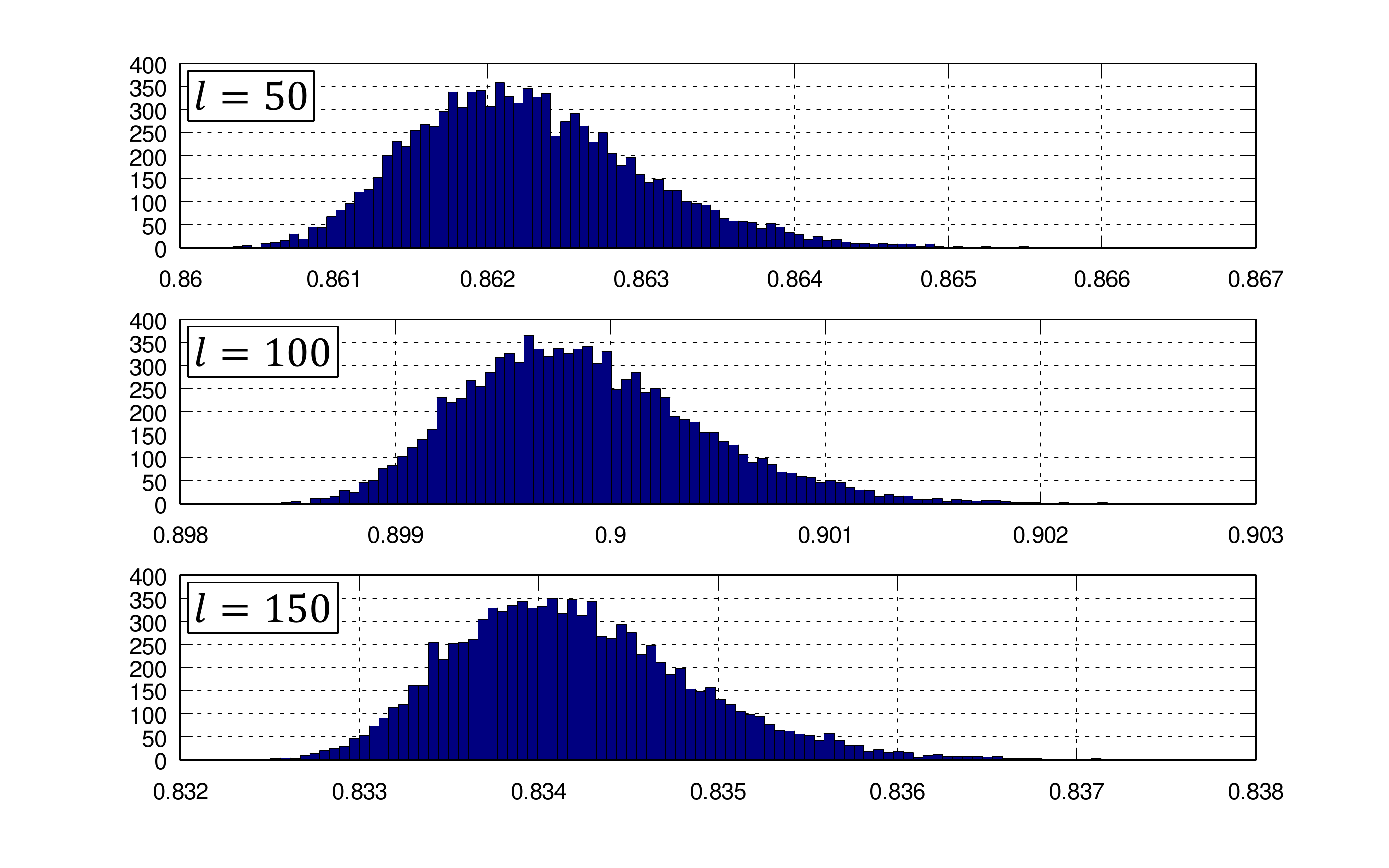}
\par\end{centering}
\caption{Histograms of the first node in layer $l=50$, $l=100$ and $l=150$.\label{fig:Histograms-of-the} }
\end{figure}

Using the above method, we next initialize a network consisting of
$L=170$ layers with $s_{1}=784$ nodes in the input layer and $s_{l}=50$
nodes in each consecutive layer. The input signal is uniformly distributed
between 0 and 1. The histograms of the first node in layer $l=50$,
$l=100$ and $l=150$ is depicted in Fig.\ \ref{fig:Histograms-of-the}. 

Although the variance of the activations is slightly decreasing with
every layer, it is not vanishing until only very late layers \cite{Glorot2010}.
Furthermore no dying or saturated neurons are observed as is the case
of ReLUs \cite{Maas2014}. Same behaviour is observed for the backpropagted
gradients $\delta_{i}^{\left(l\right)}$ in (\ref{eq:Backpropagation_delta}).

\section{Alternating Optimization Method}

We discovered a technique, which improves the convergence in training
deep neural networks ($L>50$) drastically. The trick is to optimize
the sets of weights $r_{j}^{\left(l\right)}$ and $c_{j}^{\left(l\right)}$
for all nodes and all layers in an alternating way. For a certain
number of iterations $i_{c}$ during an epoch only weights $r_{j}^{\left(l\right)}$
are optimized, while $c_{j}^{\left(l\right)}$ are treated as constants.
Then for a certain number of iterations $i_{r}$ the opposite is done.
Subsequently, the first step is repeated again and so on. As a finalization
step in late epochs optionally a combined optimization can be performed.
A positive side effect of applying the alternating optimization method
lies in the reduced computational load.

\section{Regularization$\protect\phantom{}$}

To help prevent overfitting, we make use of L2 regularization (also
called a weight decay term). Since two sets of parameters are optimized
in neural networks based on distance measures the overall cost function
is of the form:
\begin{equation}
\begin{aligned}J\left(R,C\right) & =\frac{1}{m}\sum_{i=1}^{m}J\left(R,C;\ x^{\left(i\right)},y^{\left(i\right)}\right)\\
 & +\frac{\lambda_{\mathrm{c}}}{2}\sum_{l=1}^{L}\sum_{i=1}^{s_{l}}\sum_{j=1}^{s_{l+1}}\left[c_{j}^{\left(l\right)}\right]_{i}^{2}+\frac{\lambda_{\mathrm{r}}}{2}\sum_{l=1}^{L}\sum_{i=1}^{s_{l}}\sum_{j=1}^{s_{l+1}}\left[r_{j}^{\left(l\right)}\right]_{i}^{2}
\end{aligned}
\label{eq:regularization}
\end{equation}
where $m$ denotes the number of samples in the training set and $\lambda_{\mathrm{c}}$
and $\lambda_{\mathrm{r}}$ are the regularization strengths. In general
$\lambda_{\mathrm{c}}$ and $\lambda_{\mathrm{r}}$ have to be selected
independently for getting optimal results. The second and third term
in (\ref{eq:regularization}) tend to decrease the magnitude of the
weights, and thus helps prevent overfitting.

Another way of preventing overfitting when using neural networks based
on distance measures is to apply the alternating optimization method
introduced in the previous section. Optimizing only for the half of
the coefficients at each moment reduces the actual number of degrees
of freedom, though acting as a form of regularization.

\section{Experiments$\protect\phantom{}$}

\subsection{MNIST}

In this subsection we demonstrate that the classification performance
of the presented neural networks based on distance measures and Gaussian
activation functions is comparable to the classification performance
of traditional neural networks. To this end we use the well-known
MNIST benchmark of handwritten digit images. MNIST consists of two
datasets of handwritten digits of size 28x28 pixels, one for training
(60,000 images) and one for testing (10,000 images). The goal of the
experiment is not primarily to achieve state-of-the-art results, which
would require convolutional layers and novel elastic training image
deformations.

\begin{figure}[tbh]
\begin{centering}
\includegraphics[bb=-50bp 60bp 850bp 450bp,scale=0.25]{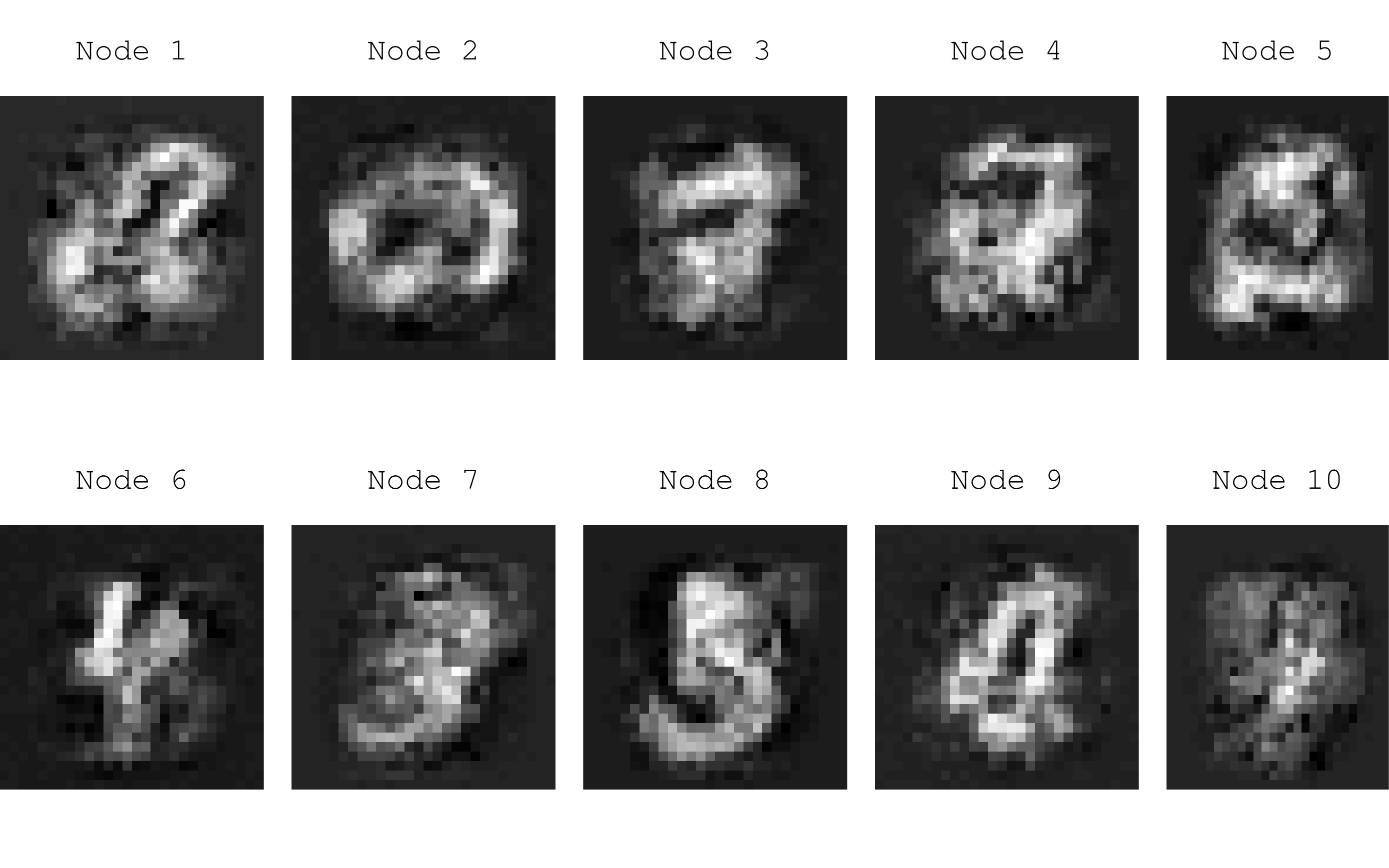}
\par\end{centering}
\caption{Learned first layer weights $c_{j}^{\left(1\right)},\ j\in\left\{ 1,\ldots10\right\} $
at the end of learning for MNIST\label{fig:Learned-first-layer_Centroid} }
\end{figure}

The input signal consists of raw pixel intensities of the original
gray scale images mapped to real values pixel intensity range from
0 (background) to 1 (max foreground intensity). The dimension of the
input signal is 28x28 = 784. The network architecture is built up
of 2 fully-connected hidden layers with 100 units in each layer. We
use the cross-entropy cost function in (\ref{eq:Quadratic_cost})
and apply regularization (\ref{eq:regularization}) only to the radii
terms, i.e. $\lambda_{\mathrm{c}}=0$. We use a variable learning
rate that shrinks by a multiplicative constant after each epoch. 

\begin{figure}[tbh]
\begin{centering}
\includegraphics[bb=-50bp 60bp 850bp 450bp,scale=0.25]{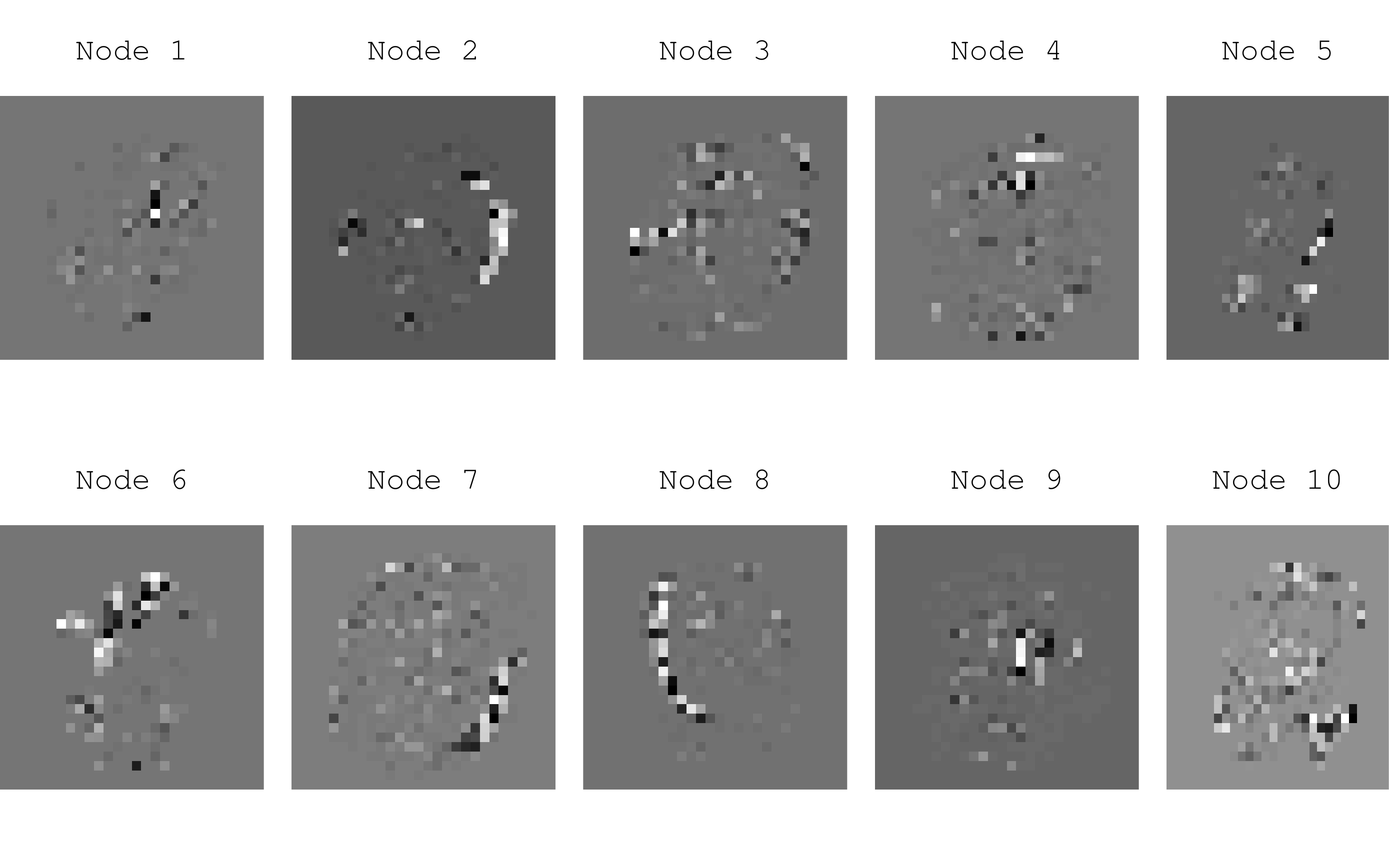}
\par\end{centering}
\caption{Learned first layer weights $r_{j}^{\left(1\right)},\ j\in\left\{ 1,\ldots10\right\} $
at the end of learning for MNIST\label{fig:Learned-first-layer_Radii}}
\end{figure}

The test set accuracy after 30 epochs is 98.2\%, which is a comparable
result to a traditional neural networks with approx.\ the same number
of parameters \cite{Erhan2009}. The weights of the first layer are
depicted in Fig.\ \ref{fig:Learned-first-layer_Centroid} and Fig.\ \ref{fig:Learned-first-layer_Radii}.

\subsection{Images}

\begin{figure}[tbh]
\begin{centering}
\includegraphics[bb=-70bp 130bp 850bp 350bp,scale=0.4]{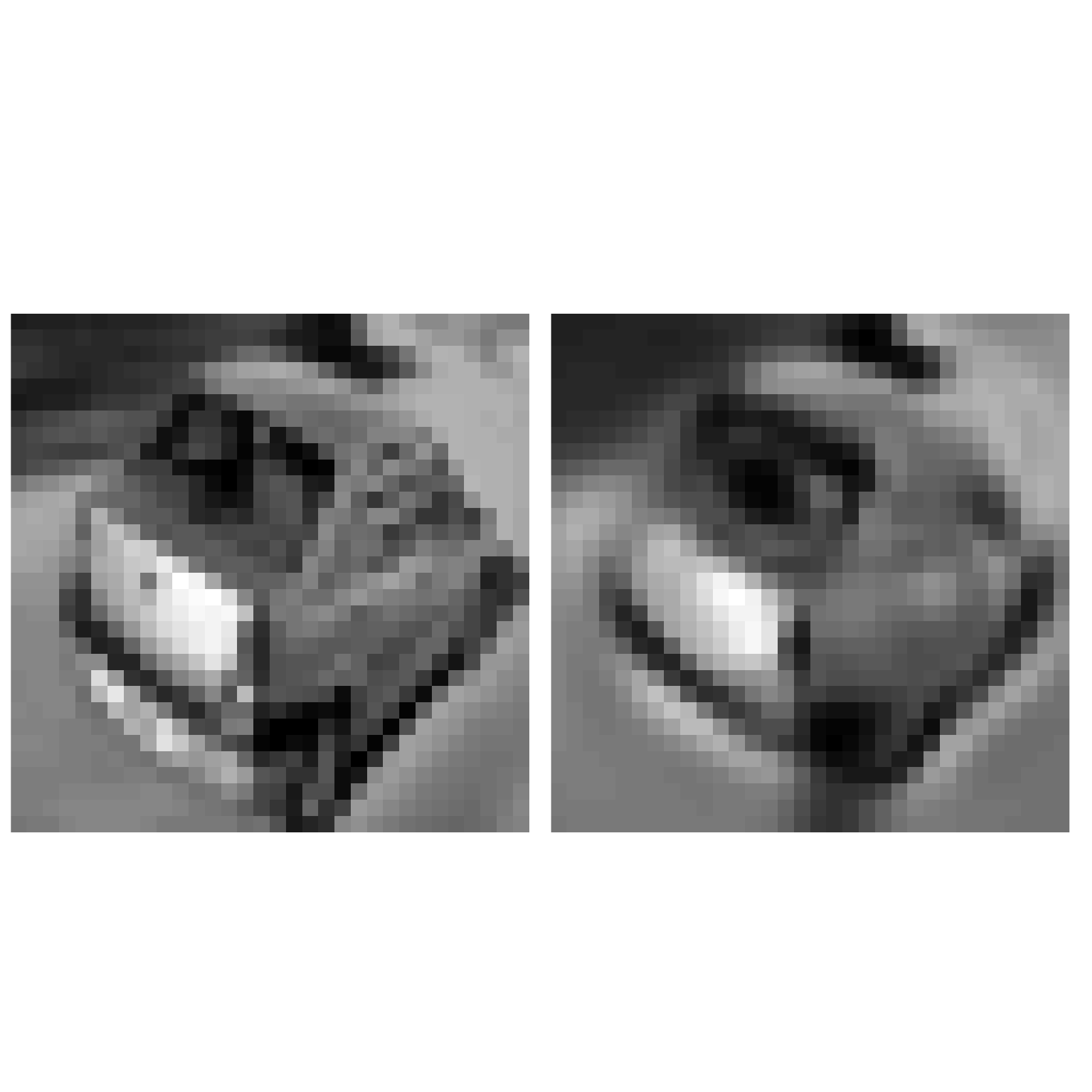}
\par\end{centering}
\caption{32x32 natural grayscale image (left) and its approximation using a
network with 2 hidden layers\label{fig:32x32-natural-grayscale}}
\end{figure}

In this section we demonstrate the ability of the presented neural
networks based on distance measures and Gaussian activation functions
to approximate arbitrary functions. To this end we use a small 32x32
grayscale natural image. Images in its most general form can be interpreted
as functions $f$ from $\mathbb{R}^{2}$ to $\mathbb{R}$. We use
a small fully-connected network with 2 hidden layers and 10 nodes
in each layer. The input of the network consists of pixel coordinates
$\left(x,\ y\right)$. The desired output is merely the grayvalue
at the corresponding pixel coordinate. We apply a full batch gradient
descent. We use the quadratic cost in (\ref{eq:Quadratic_cost}) and
no regularization to utilize all degrees of freedom. The result of
the approximation is depicted in Fig.\ \ref{fig:32x32-natural-grayscale}.

Since neural networks in general are capable of approximating only
continuous functions, the approximated image exhibits a blurry appearance.
With a slightly increased number of layers and hidden units the approximation
can be made indistinguishable from the original.

\subsection{Approximating a probability density function}

\begin{figure}[tbh]
\begin{centering}
\includegraphics[bb=-100bp 40bp 599bp 458bp,scale=0.2]{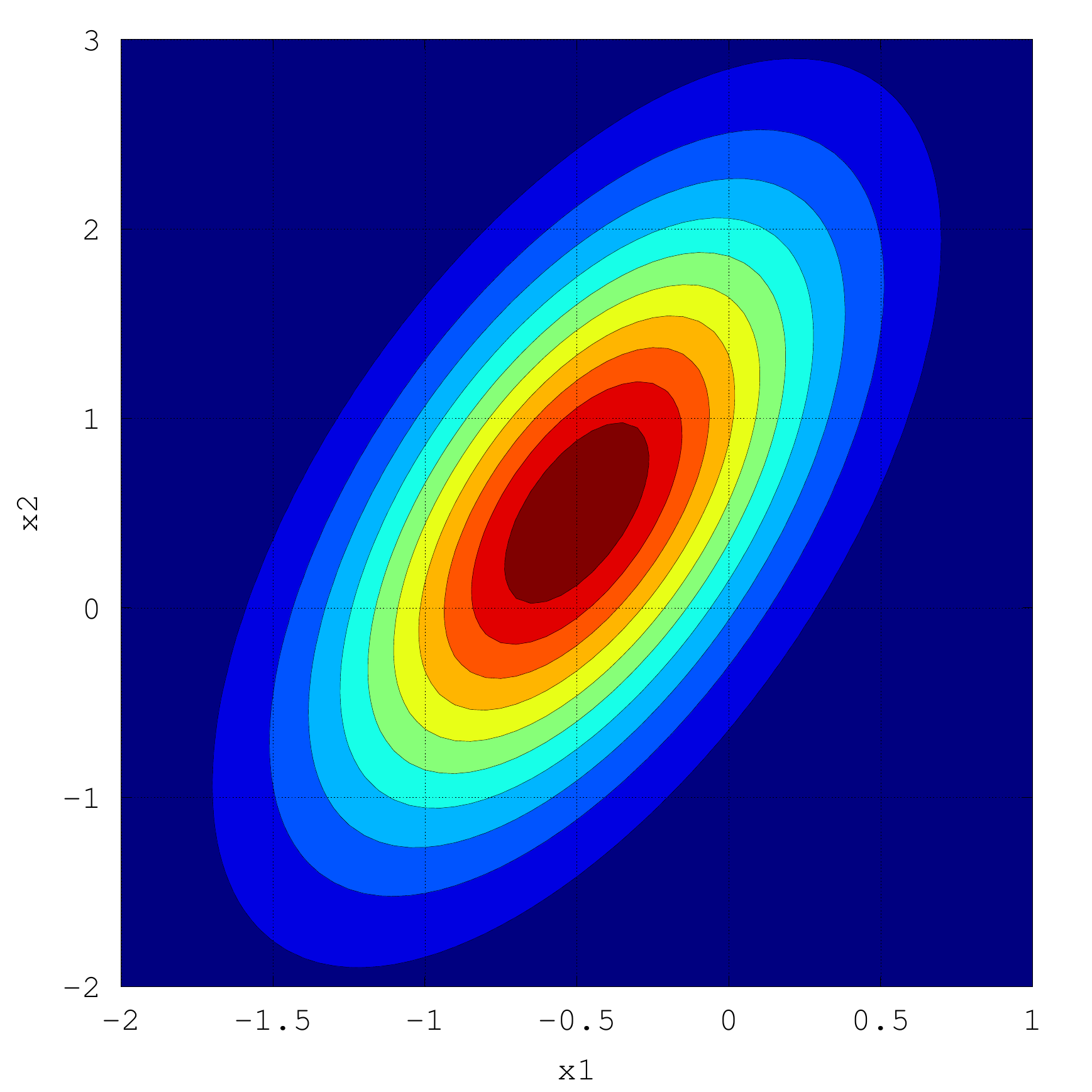}\includegraphics[bb=0bp 40bp 599bp 458bp,scale=0.2]{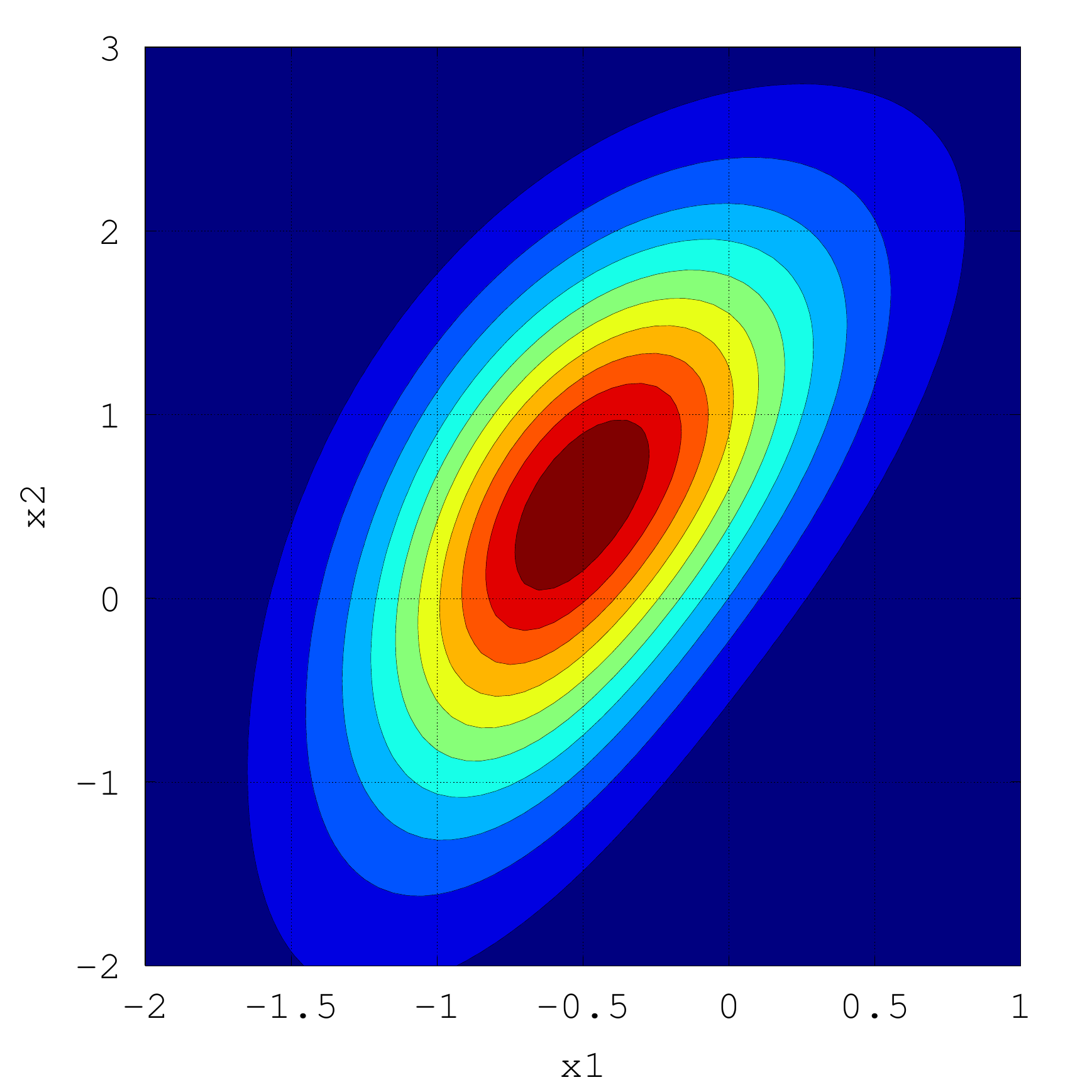}
\par\end{centering}
\caption{ Bivariate Gaussian distribution with a non-diagonal covariance matrix
and its approximation\label{fig:-Bivariate-Gaussian}}
\end{figure}

In the last example, we approximate a simple probability density function
(pdf) consisting of a bivariate Gaussian distribution with a covariance
matrix not being the identity matrix , i.e. a rotated Gaussian function,
see. Fig.\ \ref{fig:-Bivariate-Gaussian}:
\begin{equation}
p_{\mathbf{X}}\left(\mathbf{x}\right)=\frac{1}{2\pi\sqrt{\left|\Sigma\right|}}\exp\left(-\frac{1}{2}(\mathbf{x}-\mathbf{\mu})^{\mathrm{T}}\mathbf{\Sigma}^{-1}(\mathbf{x}-\mathbf{\mu})^{\mathrm{T}}\right)
\end{equation}
where
\begin{equation}
\mathbf{\mu}=\left(\begin{array}{c}
-0.5\\
0.5
\end{array}\right),\ \mathbf{\Sigma}=\left(\begin{array}{cc}
0.30 & 0.36\\
0.36 & 1.20
\end{array}\right).\label{eq:regularization-1}
\end{equation}
Since each layer in neural networks based on distance measures and
Gaussian activation functions consists only of Gaussian functions
aligned with the main axes (\ref{eq:Input_Output_Layer_l}), we want
to see how well a minimal network consisting of 1 fully-connected
hidden layer with 2 units is able to approximate this function. The
result is depicted on the right hand side of Fig.\ \ref{fig:-Bivariate-Gaussian}.
The approximation in the the vicinity of the peak is excellent. In
the surrounding area a slight distortion is visible. The overall RMS
of the approximation is 0.006.

\section{Conclusions and Future Work$\protect\phantom{}$}

In this paper, we have revisited neural network structures based on
distance measures and Gaussian activation functions. We showed that
training of these type of neural network is only feasible when using
the stochastic gradient descent in combination with RMSProp. We showed
also that with a proper initialization of the networks the vanishing
gradient problem is much less than in traditional neural networks.

In our future work we will examine neural network architectures which
are built up of combinations of traditional layers and layers based
on distance measures and Gaussian activation functions introduced
in this paper. In particular the use of layers based on distance measures
in the output layer is showing very promising results. Currently we
are analyzing the performance of neural networks based on distance
measures when used in deep recurrent networks.

\bibliographystyle{plain}
\bibliography{Paper2017}

\end{document}